\documentclass[runningheads]{llncs}

\usepackage[T1]{fontenc}
\usepackage[utf8]{inputenc}

\usepackage{graphicx}
\usepackage{booktabs}
\usepackage{multirow}
\usepackage{tabularx}
\usepackage{amsmath}
\usepackage{makecell}
\usepackage{multicol}
\usepackage{orcidlink}
\usepackage{amsmath,amssymb,amsfonts}
\usepackage{xcolor}
\usepackage{hyperref}
\hypersetup{
    hidelinks
}

\newcommand{\best}[1]{\textbf{#1}}

\title{Query Rewriting for Complex Object Segmentation in 4D Gaussian Representations}

\titlerunning{Query Rewriting for Complex Object Segmentation}

\author{Thanh-Khoi Nguyen\inst{1,2}\orcidlink{0009-0003-5879-451X}
Thien-Phuc Tran \thanks{Corresponding author}\inst{1, 2}\orcidlink{0009-0008-0800-6884}
Minh-Triet Tran \inst{1,2}\orcidlink{0000-0003-3046-3041}}
\authorrunning{Thanh-Khoi Nguyen and Thien-Phuc Tran}

\institute{University of Science, Ho Chi Minh City, Vietnam \and
Viet Nam National University, Ho Chi Minh City, Vietnam \\
\email{23120009@student.hcmus.edu.vn, ttphuc@selab.hcmus.edu.vn, tmtriet@fit.hcmus.edu.vn}
}

\begin{document}

\maketitle

\begin{abstract}
Recent 4D Gaussian representation frameworks have demonstrated strong performance in language-guided dynamic scene understanding. However, these methods remain highly sensitive to verbose and narrative-style queries that contain noisy contextual information. In this paper, we investigate the impact of query rewriting for complex object segmentation in 4D Gaussian representations. Inspired by recent findings in retrieval-augmented language models and keyword-guided query reformulation, we propose a training-free reinterpretation strategy that transforms long descriptive queries into concise keyword-grounded forms. Our approach progressively reduces linguistic noise while preserving semantic anchors relevant to object-centric representations. Experiments on HyperNeRF and Neu3D demonstrate that concise rewritten queries significantly improve both temporal localization and spatial segmentation performance. In particular, our method improves average temporal accuracy from 60.92\% to 92.21\% and average vIoU from 20.08\% to 76.94\% without any additional fine-tuning. Extensive ablation studies further reveal that shorter, keyword-focused queries consistently yield stable video-feature similarity distributions and better alignment with object-centric Gaussian representations.

\keywords{4D Gaussian Splatting \and Query Rewriting \and Temporal Segmentation \and Video Understanding \and Language-Guided Segmentation}
\end{abstract}

\section{Introduction}

Language-guided retrieval and segmentation in dynamic scenes have recently attracted significant attention due to the emergence of 4D Gaussian representation frameworks \cite{Wu_2024_CVPR}. Methods such as 4D LangSplat \cite{li20254dlangsplat4dlanguage} enable language-conditioned querying over 4D Gaussians by associating semantic embeddings with spatiotemporal scene elements. Despite their effectiveness, these approaches exhibit substantial sensitivity to query formulation.

In practical settings, users rarely provide concise object-centric queries. Instead, queries are often verbose, narrative, ambiguous, or conversational. The core issue with long-form queries is that they often extend beyond describing the target object and incorporate substantial peripheral context. This introduces irrelevant signals into the input, effectively acting as noise without contributing meaningful discriminative information for retrieval.

According to Schema Theory in cognitive psychology, humans acquire new knowledge most effectively when incoming information is closely aligned with existing prior knowledge structures. When a concept is expressed using unfamiliar, verbose, or ambiguous language, it becomes significantly more difficult for learners to comprehend and retain~. In contrast, when instructors employ familiar keywords and core concepts as anchoring cues, learners can more readily assimilate new information into their existing schemas, thereby substantially improving learning efficiency. Analogously, retrieval models - particularly those trained on concise and well-structured captions - struggle when processing long, complex queries that deviate from the training distribution. Existing studies suggest a semantic gap between natural user expressions and the representations used for retrieval. Furthermore, long and unfamiliar queries often introduce hallucinated or out-of-scope semantic concepts, reducing retrieval robustness. 

Motivated by these observations, we investigate the role of query rewriting for language-guided segmentation in 4D Gaussian representations. Rather than expanding queries with additional context, we explore the opposite direction: progressively simplifying and grounding queries toward concise keyword-focused forms. Our approach is training-free and requires no modifications to the underlying 4D LangSplat architecture.

Specifically, we introduce a multi-level rewriting framework that transforms verbose narrative descriptions into compact keyword-grounded expressions. We demonstrate that shorter and more familiar queries significantly improve temporal localization and segmentation quality. Extensive experiments on HyperNeRF~\cite{park2021hypernerf} and Neu3D~\cite{Li_2022_CVPR} dataset show substantial improvements across all evaluated scenes.

Our contributions are summarized as follows:

\begin{itemize}
    \item We expose a critical structural vulnerability in current language-guided 4D Gaussian pipelines: the 'semantic gap' between natural user formulations and the object-centric supervision paradigm used to train 4D Gaussians. We demonstrate that these models do not merely struggle with query length, but fail fundamentally due to the diffusion of spatial activations caused by peripheral context
    
    \item We propose a simple yet effective training-free query-rewriting framework. By leveraging an LLM to dynamically extract attributes rather than fine-tuning per scene, our approach remains scene-independent while preserving open-vocabulary capabilities, thereby enhancing query interpretability, thus improving temporal localization and segmentation quality per query.

    
    \item We provide extensive ablation studies analyzing the effects of query length and keyword grounding on retrieval performance.
\end{itemize}

\section{Related Work}
\label{sec:RelatedWork}

\textbf{4D Gaussian Splatting for Dynamic Scenes.}
Our work builds on recent advances in 4D Gaussian splatting for real-time dynamic scene rendering. 
Wu et al.\ introduced 4D Gaussian Splatting (4D-GS), which augments static 3D Gaussian splats with temporal components to enable efficient novel view synthesis of dynamic scenes while maintaining high visual fidelity~\cite{Wu_2024_CVPR}. 
Subsequent extensions have explored alternative parameterizations and efficiency improvements, such as 4D-Rotor Gaussian Splatting (4D-RotorGS) for anisotropic XYZT Gaussians~\cite{duan:2024:4drotorgs} and orientation-anchored Gaussians for robust 4D reconstruction from monocular video~\cite{wu2025oriGS}. 
These works primarily focus on geometric and photometric modeling, whereas we target robustness to language-guided segmentation on top of existing 4D Gaussian representations.

\textbf{Language-guided 4D Gaussian Splatting.}
Several recent approaches have begun to enrich 4D Gaussian representations with semantic and linguistic cues.
4-LEGS extends this direction by embedding dense spatiotemporal language features directly into dynamic Gaussian representations, enabling fine-grained text-conditioned localization and editing across both space and time in dynamic scenes~\cite{fiebelman20244legs}.
Recently emerging as a generalist approach, 4D LangSplat learns ``language fields'' over 4D Gaussians via multimodal large language models (MLLMs), enabling both time-agnostic and time-sensitive open-vocabulary queries in dynamic scenes~\cite{li20254dlangsplat4dlanguage}. 
While recent systems have focused heavily on architectural and training strategies to align language with 4D geometry, they implicitly assume that users will provide optimal, object-centric text inputs. Our work investigates the query-side vulnerability of 4D Gaussian representations. By demonstrating that standard language-vision embeddings are highly unstable under natural, verbose formulations, we introduce a training-free structural intervention that maps natural language into the specific geometric and temporal parameters required by 4D Gaussians.

\textbf{4D Segmentation and Semantic Scene Editing.}
In parallel with advances in representation and language conditioning, there has been growing interest in segmentation and editing in 4D Gaussian spaces. 
Ji et al.\ introduced Segment Any 4D Gaussians (SA4D), which generalizes ``segment anything'' capabilities to 4D worlds and supports operations such as object removal, recoloring, and composition on 4D Gaussians~\cite{ji2024segment4dgaussians}. 
Wei's thesis presents a Gaussian splatting-based 4D segmentation framework for volumetric video that enables semantic editing of dynamic scenes and offers a comprehensive treatment of 4D segmentation pipelines, losses, and editing operations~\cite{wei2024_semantic4d_thesis}. 
Compared to these approaches, our focus is not on proposing a new segmentation backbone, but on improving the robustness and stability of language-guided 4D segmentation by manipulating the query formulation itself.

\textbf{Open-vocabulary 3D/4D Semantics.}
Our approach is intended to bridge the research gap in open-vocabulary 3D scene understanding, in which neural fields are supervised by language to support flexible semantic queries. 
A line of work on language fields for 3D scenes uses CLIP-like supervision to map spatial locations to semantic embeddings, enabling open-vocabulary querying and editing; we refer to a representative method in this space as open-vocabulary 3D scene understanding with neural fields and language supervision~\cite{li2025langsplatv2,qin2024langsplat,zhou2024feature,lee2025cf3}. 
4D LangSplat can be viewed as a 4D analogue of these language fields, extending them to dynamic scenes and 4D Gaussian representations~\cite{li20254dlangsplat4dlanguage}. 
In contrast to designing new language fields, our method assumes a pretrained language-vision space and instead studies how different query formulations - from narrative descriptions to concise keyword lists - affect retrieval and segmentation performance in that space.

\textbf{Foundational Vision-Language Models.}
Finally, our method leverages foundational vision--language and segmentation models that have become standard components in open-vocabulary perception pipelines. Segment Anything (SAM) provides a high-capacity segmentation model that can generate object masks from 2D images with minimal supervision, and has been widely adopted as a geometric backbone in 3D and 4D pipelines~\cite{Kirillov_2023_ICCV,ravi2024sam2segmentanything,carion2026sam3}. 
CLIP~\cite{radford2021clip} learns a joint image-text embedding space from natural language supervision at scale, enabling open-vocabulary image retrieval, classification, and segmentation. In our framework, SAM provides object-centric supervision in the image domain, while CLIP (or similar encoders) defines the language-vision embedding space, whose behavior we analyze under different query formulations; our query-rewriting strategy is designed to better exploit this space without any additional training.

\section{Method}
\label{sec:Method}

Our approach introduces a training-free, multi-level query-rewriting framework that improves language-guided segmentation in 4D Gaussian representations without requiring modifications to the underlying 4D LangSplat architecture. As discussed, user queries are often verbose and include not only descriptions of the target object but also substantial irrelevant peripheral context. Within the 4D LangSplat framework, 4D Gaussians are supervised using caption embeddings that are explicitly object-centric, capturing the deformation, motion, and state of a single object across frames. This design is particularly effective for object-centric descriptions (e.g., “a chicken” or “a cup”), but remains highly sensitive to extraneous linguistic artifacts (e.g., “sunlight” or “wind”).

\begin{figure}[!htbp]
    \centering
    \includegraphics[width=\linewidth]{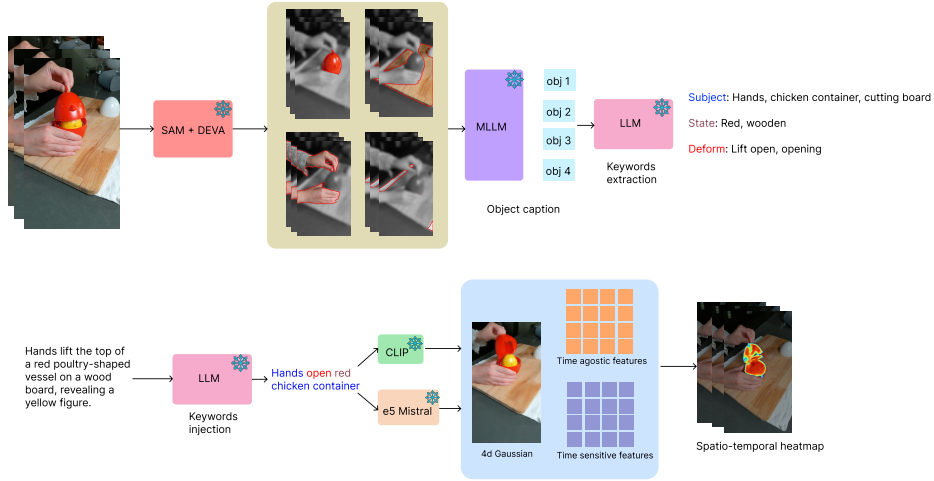}
    \caption{Overall pipeline of our framework. We propose a simple training-free query reinterpretation module that removes linguistic noise, thereby improving the alignment of embeddings with object-centric 4D Gaussians and significantly boosting both temporal localization and spatial segmentation quality. Our method leverages off-the-shelf models, without any additional training.}
    \label{fig:pipeline}
\end{figure}

Our key motivation is therefore to eliminate unnecessary contextual information and refine retrieval intent, reducing queries to their most concise and informative components. Empirical evidence further suggests that query expansion in LLM-based retrieval systems can lead to cumulative errors and degraded performance as query length increases~\cite{abe2025llmqueryexpansion}. Consequently, we focus on query compression, leveraging keywords as anchors during the rewriting process. This approach preserves semantic fidelity while constraining the generated captions to remain aligned with the target object and its associated factual attributes. The impact of keywords injection and query length is ablated in section \ref{sec:AblationStudy}. Our method is described in Figure \ref{fig:pipeline}.

\subsection{Base Framework and Feature Extraction}
We build on 4D LangSplat as our base framework for representing dynamic scenes and execute our method without any additional fine-tuning. In this framework, the dynamic 4D scene is parameterized by a set of deformable 3D Gaussians. To support semantic understanding, each Gaussian is augmented with dual semantic fields: a time-agnostic field and a time-varying field. For the time-agnostic semantic field, geometric features are first segmented into hierarchical regions using the Segment Anything Model (SAM)~\cite{Kirillov_2023_ICCV}, after which CLIP~\cite{radford2021clip} is employed to extract static feature representations. These features are subsequently passed through an autoencoder to reduce their dimensionality from 512 to 3, thereby improving memory efficiency. A parallel pipeline is used to construct the time-sensitive field: Gaussians are supervised by frame-level captions, and a status-deformable network predicts time-sensitive semantic features across different temporal instances. 

\subsection{Problem Formulation}
\label{sec:problem_formulation}

Let a dynamic 4D scene be represented by a set of deformable Gaussians, where $F^{\text{agn}}_{x,y} \in \mathbb{R}^{d_1}$ and $F^{\text{sen}}_{x,y,t} \in \mathbb{R}^{d_2}$ denote the rendered time-agnostic and time-sensitive semantic feature vectors at spatial pixel coordinates $(x, y)$ and temporal step $t$, respectively, corresponding to the dual semantic fields embedded in each Gaussian. In the standard language-guided retrieval pipeline, a user provides a natural language query $q$, which is encoded into a text embedding using either $E^{\text{agn}}_T(q) \in \mathbb{R}^{d_1}$ for the time-agnostic field, or $E^{\text{sen}}_T(q) \in \mathbb{R}^{d_2}$ for the time-sensitive field.

The retrieval pipeline proceeds in two sequential steps. First, $E^{\text{agn}}_T(q)$ is matched against $F^{\text{agn}}_{x,y}$ via a relevance score to localize the target object and produce a spatial candidate mask $\mathcal{M}_{x,y}$. Second, the cosine similarity between $E^{\text{sen}}_T(q)$ and the time-sensitive features $F^{\text{sen}}_{x,y,t}$ is computed exclusively within $\mathcal{M}_{x,y}$:
\begin{equation}
    \mathcal{S}(q, F^{\text{sen}}_{x,y,t}) = 
    \frac{E^{\text{sen}}_T(q) \cdot F^{\text{sen}}_{x,y,t}}
         {\|E^{\text{sen}}_T(q)\| \|F^{\text{sen}}_{x,y,t}\|}, 
    \quad (x,y) \in \mathcal{M}_{x,y}
\end{equation}
A frame $t$ is classified as active if $\mathcal{S}$ exceeds a predefined threshold $\tau$, and inactive otherwise.

\vspace{1mm}
\noindent\textbf{The Semantic Gap.} The core vulnerability of this formulation arises from the inherent nature of the input query $q$. In practical scenarios, users rarely input a concise, object-centric query $q_\text{anchor}$ (e.g., ``a cup''). Instead, they typically provide a verbose, descriptive query $q_\text{long}$ (e.g., ``a clear glass cup resting on a wooden table while sunlight hits it'').

We formulate $q_\text{long}$ as a composition of the core semantic intent $q_\text{anchor}$ and extraneous peripheral context $q_\text{noise}$:
\begin{equation}
    q_\text{long} = q_\text{anchor} \cup q_\text{noise}
\end{equation}

Because the foundation text encoder $E_T(\cdot)$ processes the sentence holistically, the presence of $q_\text{noise}$ shifts the resulting embedding $E_T(q_\text{long})$ away from the optimal object-centric representation in the shared feature space. This semantic shift introduces irrelevant signals, leading to diffused spatial activations and unstable temporal similarity scores $\mathcal{S}(q_\text{long}, F_{x,y,t})$.

\vspace{1mm}
\noindent\textbf{Optimization Objective.} To bridge this semantic gap without altering the underlying 4D LangSplat architecture or requiring additional fine-tuning, our objective is to define a training-free rewriting function $\mathcal{R}(\cdot)$ that extracts the core semantic anchors from the verbose query:

\[
    q_\text{anchor} = \mathcal{R}(q_\text{long})
\]

The ideal rewriting function minimizes the linguistic noise such that the similarity response of the rewritten query produces a precise spatial mask $\mathcal{M}_{x,y}$ via $E^{\text{agn}}_T(\mathcal{R}(q_\text{long}))$, and maximizes the cosine similarity $\mathcal{S}$ within that mask via $E^{\text{sen}}_T(\mathcal{R}(q_\text{long}))$, jointly improving temporal localization and spatial segmentation quality.

\subsection{Query Rewriting Strategy}
\label{subsec:rewrite_query}

\noindent\textbf{Object-Level Caption Generation.} Given a monocular video sequence, our primary objective is to construct robust textual descriptions that accurately define the ground-truth semantic target features, denoted as $F_{target}$ in our formulation. Following recent advancements in multimodal scene understanding~\cite{li20254dlangsplat4dlanguage,wu20254dlangvggt}, we employ an MLLM to perform instance-specific video captioning. Crucially, this MLLM-based visual prompting process is strictly a theoretical setup used exclusively to define the ideal semantic upper bound ($F_{target}$) for our optimization objective. It is not executed during user inference. By establishing this mathematically clean, object-centric baseline offline, we provide a strict target for our real-time rewriting function $\mathcal{R}(q_{long})$ to emulate.

Initially, spatial regions within the video are extracted and tracked using the Segment Anything Model (SAM)~\cite{Kirillov_2023_ICCV} and DEVA~\cite{cheng2023tracking}. Because SAM generates hierarchical segmentations across multiple granularities, we explicitly filter the outputs to retain only object-level mask regions. During the captioning phase, we prompt the MLLM to generate per-object descriptions that comprehensively encapsulate four semantic dimensions: (1) the inherent identity of the object, (2) its visual appearance, (3) its current physical state, and (4) its temporal deformation or transformation trajectory.

A fundamental challenge in this process is effectively guiding the MLLM's visual attention. Feeding isolated foreground masks strips away critical contextual cues, whereas providing raw video frames often causes the MLLM to hallucinate or conflate similar instances in complex dynamic scenes. To mitigate this, we adopt the visual prompting mechanism introduced in 4D LangSplat~\cite{li20254dlangsplat4dlanguage}. Specifically, we superimpose a distinct boundary contour around the target object and apply a grayscale filter to all background pixels outside this boundary. This strategy grounds the MLLM's attention directly on the target object while preserving the necessary global scene context.
\\\\
\noindent\textbf{Keyword-Anchored Query Rewriting.}
As previously discussed, augmenting the verbose query $q_\text{long}$ via query expansion introduces supplementary context, which frequently incurs the risk of semantic hallucinations and popularity bias, particularly for unfamiliar queries~\cite{abe2025llmqueryexpansion}. Consequently, we optimize the query formulation using a rigorous query-shortening mechanism. However, this compression must strictly avoid information loss by adhering to three core criteria: (1) absolute preservation of the target object's identity, (2) retention of the core semantic intent of the original query, and (3) generation of a structural formulation that is highly condensed yet precisely accurate.

To meet these criteria, prompting an LLM to perform unconstrained shortening and simple summarization is insufficient. Without explicit constraints, LLMs generalize specific concepts into broader descriptions, leading to semantic drift and the omission of critical state-level details. To address this, we introduce a Keyword Injection mechanism to establish explicit semantic anchors within our rewriting function $\mathcal{R}(\cdot)$~\cite{abe2025llmqueryexpansion}. Directly incorporating these keywords into the rewriting process preserves the core semantic intent and guides our retrieval system's attention to converge precisely on the target entity. Our rewriting mechanism serves as a structural translation layer mapping parts of the query into a strict, deterministic framework (Subject, State, Deformation).

Specifically, this pipeline uses a lightweight LLM to extract a constrained set of keywords from the object-level captions. These extracted, open-vocabulary keywords are deterministically categorized into essential attributes for 4D representation: \textit{Subject}, \textit{State}, and \textit{Deformation}. This tripartite classification is explicitly designed to align with the dual semantic fields of the underlying 4D LangSplat architecture~\cite{li20254dlangsplat4dlanguage}. Following extraction, the LLM reconstructs the query by centering it on these semantic anchors, synthesizing them into a concise, grammatically coherent structure. This rewriting operation serves as a semantic filter, substantially mitigating the influence of the peripheral context $q_{noise}$ rather than explicitly masking it. Consequently, the resulting rewritten query $q_{anchor}$ yields a more stable semantic representation. 

\subsection{Integration and Inference Pipeline}

Once the optimal, keyword-anchored query $q_{anchor}$ is extracted via our rewriting function $\mathcal{R}$, it is mapped into the shared language-vision embedding space using pretrained text encoders to yield the semantic embeddings, $E^{\text{agn}}_T(q_{\text{anchor}})$ and $E^{\text{sen}}_T(q_{\text{anchor}})$. Following 4D LangSplat, we employ a two-stage querying pipeline to isolate the target object and extract the final dynamic segmentation mask. First, $E^{\text{agn}}_T(q_{\text{anchor}})$ is matched against the time-agnostic 
features $F^{\text{agn}}_{x,y}$ via a relevance score to produce the spatial candidate 
mask $\mathcal{M}_{x,y}$, establishing the geometric presence of the target object 
irrespective of its temporal state. Subsequently, $E^{\text{sen}}_T(q_{\text{anchor}})$ 
is matched against the time-sensitive features $F^{\text{sen}}_{x,y,t}$ via cosine 
similarity $\mathcal{S}$ exclusively within $\mathcal{M}_{x,y}$. Frames whose score 
exceeds the threshold $\tau$ are classified as active, and $\mathcal{M}_{x,y}$ is 
retained as the final segmentation mask for those frames.

\section{Experiment}
\label{sec:Experiment}

\subsection{Experimental Setup}

\textbf{Datasets.}
We evaluate our method on four dynamic scenes from the HyperNeRF~\cite{park2021hypernerf} and four from the Neu3D~\cite{Li_2022_CVPR} dataset. These standard 3D/4D datasets primarily feature highly concise, object-centric captions that fail to capture real-world user interactions; in practice, users naturally employ conversational, narrative, and verbose phrasing that conveys substantial context. To rigorously simulate this well-documented human behavior, we utilized Gemini 3.0 \cite{pichai2025gemini3} to generate complex queries that emulate realistic human verbosity and conversational ambiguity. This ensures the queries realistically mimic the 'semantic gap' by including descriptive noise, without containing explicit keyword matches that directly refer to objects in the video, and fully reflect the four-level query complexity we proposed in Section \ref{sec:Method}.

\textbf{Implementation Details.}
Following the 4D LangSplat pipeline, we first use the time-agnostic features to identify candidate masks for the referenced object, applying a cosine similarity threshold of 0.5. Subsequently, the same procedure is applied to the time-sensitive features to select the subset of masks that satisfy the query's temporal constraints. We focus on rewriting and standardizing query formulations as described in section \ref{subsec:rewrite_query}. Specifically, we employ  Qwen2-VL-7B~\cite{Qwen2-VL} as the multimodal backbone for instance-specific object captioning, e5-mistral-7b~\cite{wang2024improving} for time-sensitive text decoder and utilize Llama3-8B~\cite{grattafiori2024llama3herdmodels} as the lightweight language model to execute the keyword extraction and rewriting pipeline.

To further experiment on the effectiveness of prompt length and keywords, we rewrite the original query at different levels:
\begin{itemize}
    \item \textbf{Level 0}: Uses the original query without any modification.
    \item \textbf{Level 1}: Incorporates exactly one keyword corresponding to the referenced object, ensuring that this object does not appear at the beginning of the sentence; the query length is reduced to half of Level 0.
    \item \textbf{Level 2}: Utilizes all the generated keywords to construct the query, with the length reduced to half of Level 1.
    \item \textbf{Level 3}: Forms a semantically coherent sentence using only the generated keywords, with the shortest possible length.
\end{itemize}
All experiments are conducted on a single A100 GPU.

\textbf{Baselines.}
We build upon 4D LangSplat as our base framework without any additional fine-tuning. For time-agnostic queries, geometric features are first segmented into hierarchical regions using SAM, and then CLIP is used to extract feature representations. These features are subsequently passed through an autoencoder to reduce their dimensionality from 512 to 3 for memory efficiency. Furthermore, we include 4DLangVGGT~\cite{wu20254dlangvggt} as an additional baseline, representing a recent approach to 4D language-guided Gaussian Splatting for spatiotemporal scene understanding.


\textbf{Metrics.}
Following \cite{li20254dlangsplat4dlanguage}, we adopt three primary metrics.
\begin{itemize}
    \item \textbf{Video IoU (vIoU)} measures end-to-end pipeline performance. For each query, frames are classified as active or inactive based on whether their video feature similarity exceeds the mean similarity threshold, and vIoU is computed as the average spatial IoU between the predicted mask and ground-truth mask over correctly classified active frames.
    
    \item \textbf{Temporal Accuracy} measures the proportion of frames correctly classified as active or inactive, evaluating temporal localization independently of spatial grounding quality.
    \item \textbf{Mean IoU (mIoU)} evaluates the spatial segmentation quality for time-agnostic queries. It is calculated as the average Intersection over Union (IoU) between the predicted spatial mask and the ground-truth mask across all frames in the test set, independent of any temporal state constraints.
\end{itemize}

\subsection{Main Results}

\begin{table*}[!t]
\centering
\caption{Quantitative comparison on time-sensitive querying with complex queries on the HyperNeRF dataset~\cite{park2021hypernerf}.}
\label{tab:main}

\resizebox{\linewidth}{!}{
\begin{tabular}{lcccccccc @{\hspace{8pt}} cc}
\toprule
\multirow{2}{*}{Method} &
\multicolumn{2}{c}{americano} &
\multicolumn{2}{c}{chickchicken} &
\multicolumn{2}{c}{split-cookie} &
\multicolumn{2}{c}{espresso} &
\multicolumn{2}{c}{\textbf{Average}} \\

\cmidrule(lr){2-3}
\cmidrule(lr){4-5}
\cmidrule(lr){6-7}
\cmidrule(lr){8-9}
\cmidrule(lr){10-11}

& Acc(\%) & vIoU(\%)
& Acc(\%) & vIoU(\%)
& Acc(\%) & vIoU(\%)
& Acc(\%) & vIoU(\%)
& Acc(\%) & vIoU(\%) \\
\midrule
4DLangVGGT & 53.70 & 13.40 & 58.53 & 6.25 & 56.39 & 0.00 & 48.14 & 10.12 & 54.19 & 7.44 \\
4D LangSplat & 55.05 & 34.07 & 44.02 & 8.93 & 71.70 & 23.95 & 72.92 & 13.38 & 60.92 & 20.08 \\
\midrule

Ours
& \best{95.20} & \best{83.74}
& \best{95.65} & \best{89.02}
& \best{97.17} & \best{85.38}
& \best{80.83} & \best{49.60}
& \best{92.21} & \best{76.94} \\

\bottomrule
\end{tabular}
}
\end{table*}
\textbf{Comparison with Baselines.} Table \ref{tab:main} reports the vIoU and temporal accuracy across the four HyperNeRF scenes, where our method consistently outperforms both 4DLangVGGT and 4D LangSplat in time-sensitive querying. To prove the generalizability of our rewriting framework beyond a single dataset, we extend our evaluation to four additional distinct scenes from the Neu3D dataset (Table \ref{tab:neu3d}). As the results show, our training-free approach demonstrates strong robustness across complex, previously unseen conversational domains.

Notably, our approach does not rely on test-time adaptation or fine-tuning over a predefined set of queries. Instead, it employs a training-free query rewriting mechanism, demonstrating strong robustness when handling complex queries in previously unseen domains.

\begin{table*}[t]
\centering
\caption{Time-agnostic query on Neu3D dataset\cite{Li_2022_CVPR} using descriptive queries.}
\label{tab:neu3d}
\resizebox{\textwidth}{!}{%
\begin{tabular}{l cccc @{\hspace{8pt}} c}
\toprule
\multirow{2}{*}{Method} & \multicolumn{5}{c}{Neu3D (mIoU)} \\
\cmidrule(r){2-5}
\cmidrule(l){6-6}
& coffee-martini & cook-spinach & flame-salmon & flame-steak & \textbf{Average} \\
\midrule
4D LangSplat & 61.82 & 77.64 & 41.45 & 48.57 & 57.37 \\
Ours     & \textbf{75.43} & \textbf{86.28} & \textbf{48.13} & \textbf{91.17} & \textbf{75.25} \\
\bottomrule
\end{tabular}%
}
\end{table*}

\textbf{Impact of Query Familiarity Level.}
\begin{table}[!t]
\centering
\small
\caption{Mean vIoU improvement over the descriptive query ($\Delta$vIoU, \%) at each query familiarity level on the HyperNerf dataset \cite{park2021hypernerf}.}
\label{tab:level}

\begin{tabular}{lcccc}
\toprule
Samples &
$\Delta$vIoU (L1) &
$\Delta$vIoU (L2) &
$\Delta$vIoU (Anchor) &
Improved Queries \\

\midrule

americano & -0.73 & +2.48 & +49.67 & 6/8 \\
espresso & -0.08 & +1.63 & +36.22 & 9/12 \\
chickchicken & +41.47 & +39.70 & +80.08 & 8/8 \\
split-cookie & +19.52 & +31.07 & +61.43 & 6/8 \\

\midrule

Overall & +13.37 & +16.82 & +54.56 & 29/36 \\

\bottomrule
\end{tabular}
\end{table}

A central contribution of this work is the observation that reformulating a verbose descriptive query into a more familiar, concise form consistently improves retrieval performance. To quantify this, we evaluate each query at four abstraction levels: descriptive, Level 1, Level 2, and anchor. Table~\ref{tab:level} reports the mean vIoU improvement of each level relative to the descriptive baseline.

As shown in Table~\ref{tab:level}, reformulating queries into more familiar, keyword-grounded forms improves vIoU in 29 of 36 cases. The anchor query consistently achieves the highest performance across all datasets.

As observed, Levels 1 and 2 do not always exhibit strict sequential improvement. We attribute this variance to the holistic encoding mechanism of vision-language models like CLIP. Because these models evaluate text compositionally, retaining grammatical structure while injecting multiple keywords (Level 2) can alter the syntactic context, occasionally causing the encoded semantic attention to diffuse. This instability demonstrates that intermediate query shortening is insufficient, therefore directly motivates our Level 3 (Anchor) formulation. By stripping away extraneous syntax and reducing the query strictly to its core keyword components, Level 3 minimizes compositional noise and consistently achieves the most robust retrieval performance.


\textbf{Qualitative Results.}
\begin{figure}[!t]
\centering
\includegraphics[width=\linewidth]{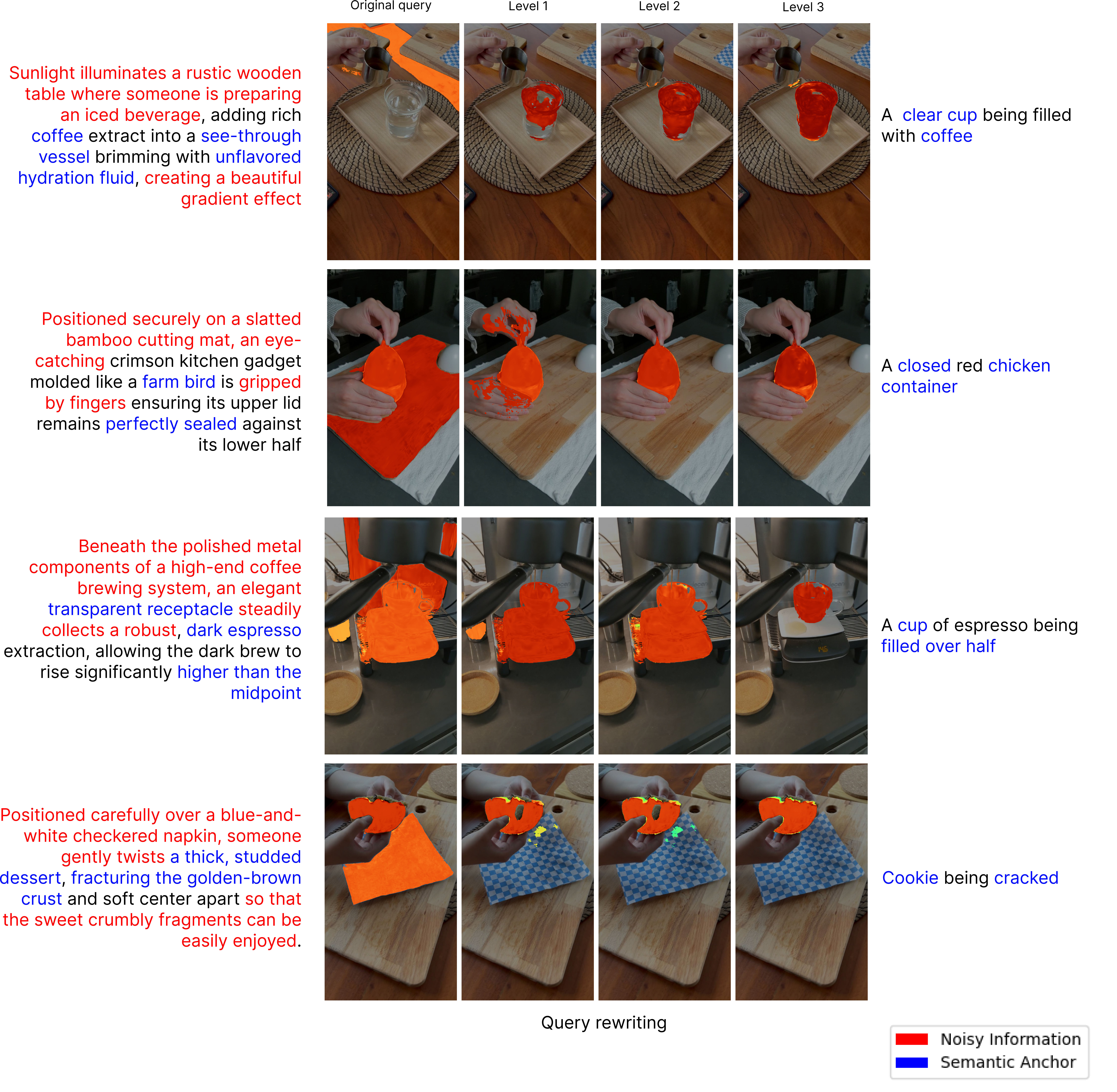}
\caption{Video feature heatmaps for the same frame across four query abstraction levels. Warmer colors indicate higher cosine similarity between video features and the query embedding.}
\label{fig:heatmap}

\end{figure}
Figure~\ref{fig:heatmap} presents video feature heatmaps for the same frame across all four query levels. The original descriptive queries produce highly diffused and noisy activation patterns that often spread into background regions. In contrast, rewritten queries produce more concentrated activations around the target object. Notably, the anchor-level queries generate tightly localized heatmaps with minimal background activation, demonstrating substantially improved semantic alignment between the query embedding and the learned Gaussian representation.

\begin{figure}[t]
\centering
\includegraphics[width=\linewidth]{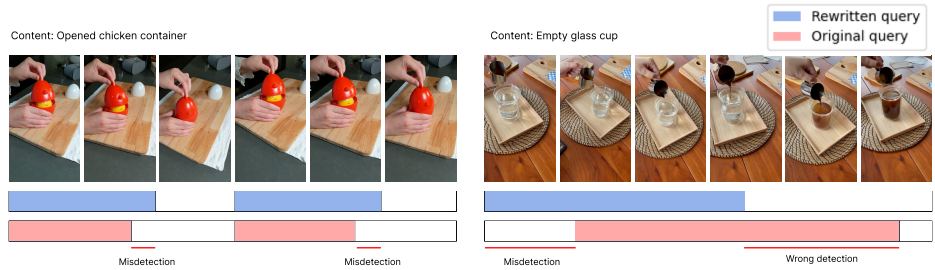}
\caption{\textbf{Visualization of temporal localization intervals.} We illustrate the frames classified as semantically active using horizontal bars. The original descriptive queries (orange) struggle with temporal coherence, resulting in fragmented activation intervals characterized by frequent misdetections and wrong detections due to peripheral linguistic noise. In contrast, our rewritten keyword-anchored queries (blue) yield stable, continuous activation segments that precisely align with the target dynamic states.}\label{fig:timeline}
\end{figure}

Figure \ref{fig:timeline} visualizes the temporal retrieval performance by representing active frames as horizontal interval bars. The original descriptive queries produce highly fragmented and unstable temporal predictions, with noticeable misdetections and false alarms. Conversely, the concise rewritten queries yield continuous, stable activation intervals that accurately span the target temporal boundaries.

\section{Ablation Study}
\label{sec:AblationStudy}

\subsection{Effect of Query Length on Retrieval Performance}

To further validate that query conciseness drives performance, we analyze the relationship between query length and retrieval metrics across all datasets.

\begin{figure}[htbp]
\centering
\includegraphics[width=\linewidth]{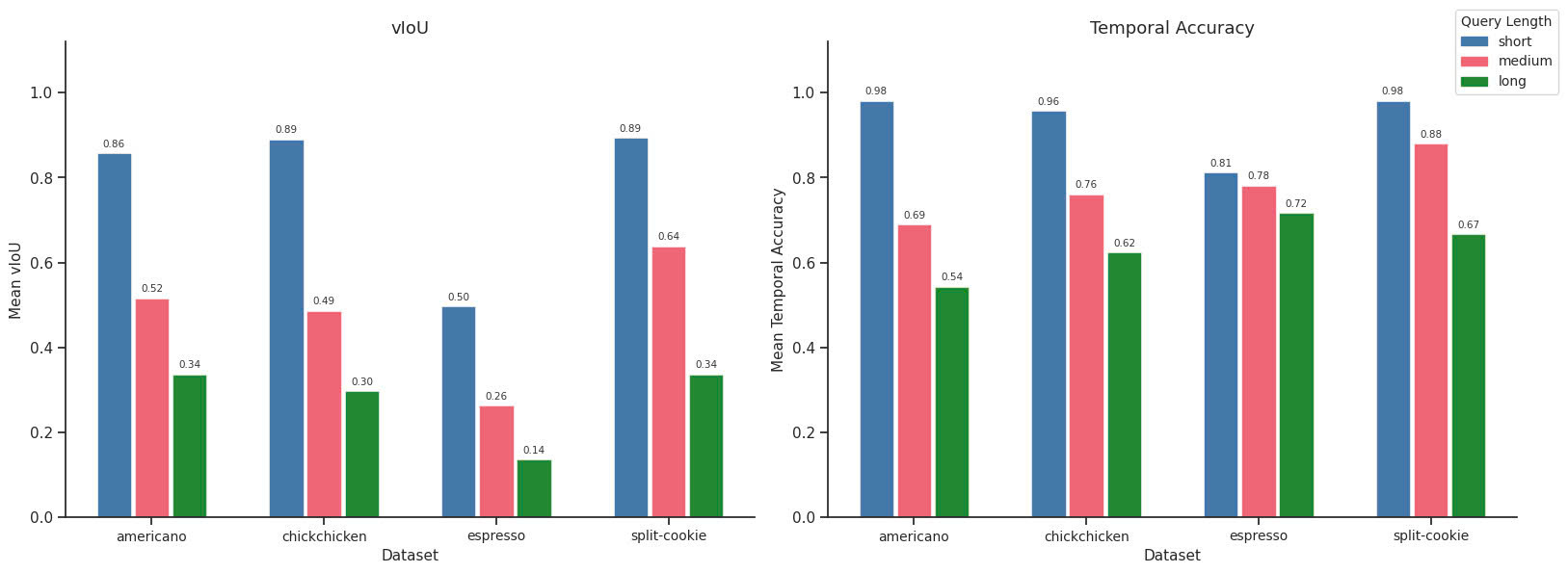}
\caption{Effect of query length on retrieval performance. Longer queries consistently produce lower performance.}
\label{fig:query_length}
\end{figure}

Figure~\ref{fig:query_length} groups queries into three categories: short ($<5$ words), medium ($5$--$15$ words), and long ($>15$ words).

The results indicate that long queries yield the worst performance among all categories. This degradation is primarily caused by the introduction of noisy contextual information that biases the embedding process.

In contrast, concise queries focus on the most salient object-centric attributes, thereby aligning more closely with the supervision paradigm used in 4D LangSplat.

\subsection{Impact of Keyword Anchoring}

We further analyze the importance of explicit keyword anchoring in the rewriting pipeline in Table \ref{tab:nokeyword-video-clip-comparison}.

\begin{table*}[t]
\centering
\small
\caption{Comparison between rewriting with and without keyword anchoring.}
\label{tab:nokeyword-video-clip-comparison}

\resizebox{\linewidth}{!}{
\begin{tabular}{lcccccccc}
\toprule

\multirow{2}{*}{Setting} &
\multicolumn{2}{c}{americano} &
\multicolumn{2}{c}{chickchicken} &
\multicolumn{2}{c}{split-cookie} &
\multicolumn{2}{c}{Average} \\

\cmidrule(lr){2-3}
\cmidrule(lr){4-5}
\cmidrule(lr){6-7}
\cmidrule(lr){8-9}

& Acc(\%) & vIoU(\%)
& Acc(\%) & vIoU(\%)
& Acc(\%) & vIoU(\%)
& Acc(\%) & vIoU(\%) \\

\midrule

Unconstrained summarization
& 51.38 & 30.99
& 61.96 & 29.53
& 74.29 & 37.71
& 62.54 & 32.74 \\

Anchored summarization
& \best{95.20} & \best{83.74}
& \best{95.65} & \best{89.02}
& \best{97.17} & \best{85.38}
& \best{96.01} & \best{86.05} \\

\bottomrule
\end{tabular}
}
\end{table*}

The results demonstrate that keyword anchoring plays a critical role in preserving the core semantics of the original query during rewriting. Even without explicit keywords, shortening the query already improves performance over the original 4D LangSplat pipeline, highlighting the general negative impact of verbose contextual descriptions. However, this simplistic unconstrained approach falls short of our proposed anchored method, as we demonstrate that injecting anchor keywords produces substantially larger gains, suggesting that concise semantic grounding is essential for robust retrieval in 4D Gaussians.

\section{Conclusion}
\label{sec:Conclusion}

In this work, we investigated the impact of query rewriting for language-guided segmentation in 4D Gaussian representations. We demonstrated that verbose, narrative-style queries significantly degrade retrieval performance by introducing noisy contextual information that is poorly aligned with object-centric supervision.

To address this limitation, we proposed a simple yet effective training-free rewriting framework that progressively transforms descriptive queries into concise keyword-grounded formulations. Extensive experiments on HyperNeRF and Neu3D showed substantial improvements in both temporal localization and segmentation quality without requiring any fine-tuning or architectural modifications.

Our analysis further revealed that shorter, semantically anchored queries yield more stable video-feature similarity distributions and stronger alignment with target objects. These findings suggest that query formulation itself is a critical factor in language-guided 4D scene understanding.

In future work, we plan to investigate automatic keyword discovery, adaptive rewriting strategies, and more principled modeling of the relationship between linguistic familiarity and visual retrieval performance.

\bibliographystyle{splncs04}
\bibliography{reference}

\end{document}